\def\year{2021}\relax
\documentclass[letterpaper]{article} 
\usepackage{aaai21}  
\usepackage{times}  
\usepackage{helvet} 
\usepackage{courier}  
\usepackage[hyphens]{url}  
\usepackage{graphicx} 
\usepackage{natbib}  
\usepackage{amsmath}
\usepackage{amsfonts}
\usepackage{caption}
\usepackage{subcaption}

\usepackage{caption} 
\usepackage[switch]{lineno}
\usepackage{booktabs}
\nocopyright

\usepackage{multirow}
\newcommand{\vpara}[1]{\vspace{0.00in}\noindent\textbf{#1 }}
\def\bpi{{\boldsymbol{\pi}}}
\title{What Makes a Star Teacher?
A Hierarchical BERT Model for Evaluating Teacher's Performance in Online Education  }

\author{

    Wen Wang$^1$, Honglei Zhuang$^2$, Mi Zhou$^3$, Hanyu Liu$^1$, Beibei Li$^1$\\

}
\affiliations{

    \textsuperscript{\rm 1}Carnegie Mellon University, \{wenw3, hanyuliu, beibeili\}@andrew.cmu.edu\\
    \textsuperscript{\rm 2}Google Research, hlz@google.com\\
        \textsuperscript{\rm 3}University of British Columbia, mi.zhou@sauder.ubc.ca\\

}

\begin{document}
\maketitle

\begin{abstract}
 
Education has a significant impact on both society and personal life. With the development of technology, online education has been growing rapidly over the past decade. While there are several online education studies on student behavior analysis, course concept mining, and course recommendation~\cite{feng2019understanding,pan2017course}, there is little research on evaluating teachers' performance in online education. 
In this paper, we conduct a systematic study to understand and effectively predict teachers' performance using the subtitles of 1,085 online courses.
Our model-free analysis shows that teachers’ verbal cues (e.g., question strategy, emotional appealing, and hedging) and their course structure design are both significantly correlated with teachers' performance evaluation. Based on these insights, we then propose a hierarchical course BERT model to predict teachers' performance in online education. Our proposed model can capture the hierarchical structure within each course as well as the deep semantic features extracted from the course content. 
Experiment results show that our proposed method achieves a significant gain over several state-of-the-art methods. Our study provides a significant social impact in helping teachers improve their teaching style and enhance their instructional material design for more effective online teaching in the future.

\end{abstract}

\section{Introduction}
Education is one of the most important industries for the global economy. The market size of the educational services industry in the United States (US) is \$1.6 trillion in 2020 \cite{edumarketsize}. Education is not only essential at a global level but also significant at a personal level because it plays a crucial role in everyone’s life. With the development of technologies, the online education market has been growing rapidly over the past decade. It is predicted that the global online education market will reach a total market size of \$319 billion by 2025 \cite{onlineedumarketsize}.   

The impact of online education services has become even more important during the ongoing COVID-19 pandemic,
which created an unprecedented disruption of education systems and affected nearly 1.6 billion learners in more than 190 countries\cite{unreports}. Most schools turned to online education and required teachers to move to online delivery of lessons due to school closures during this difficult time. However, a recent survey indicated that 70\% of the 1.5 million faculty members in the US had never taught an online course before the COVID-19 pandemic \cite{amastats}. Therefore, providing practical guidance for teachers to assist them better prepare for online teaching during this challenging time has a significant social impact. 
\begin{figure}[tb]
    \centering
    \includegraphics[width=0.8\linewidth]{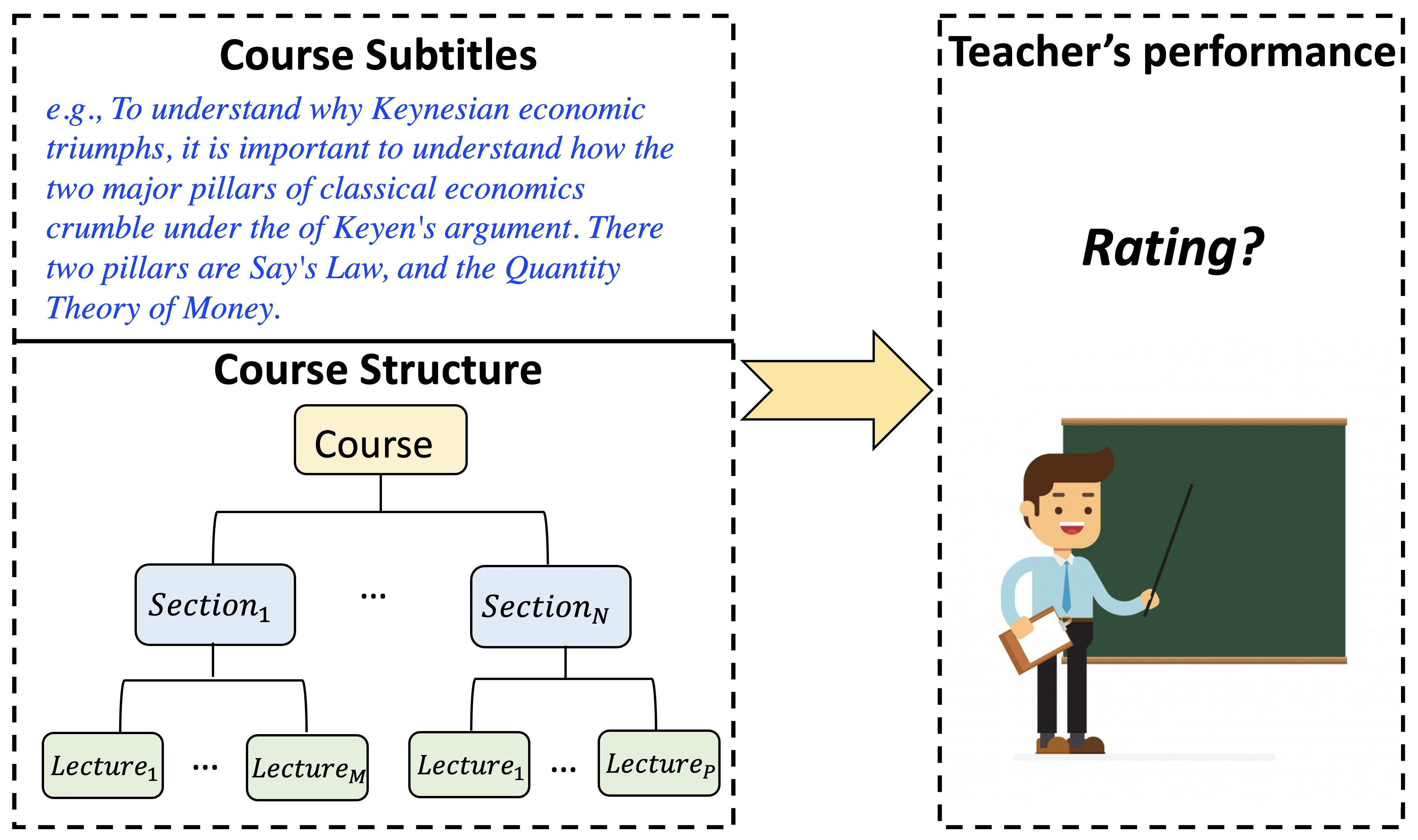}
    \caption{ Illustration of the input and output of our proposed hierarchical BERT model for predicting teachers’ performance in online education.}
    \label{fig:concept}
\end{figure}

Although there have been several related studies on student behavior analysis~\cite{feng2019understanding,trakunphutthirak2019study}, course concept mining~\cite{pan2017course,yu2019course}, and course recommendation~\cite{zhang2019hierarchical}, few of them have focused on evaluating teachers' performance based on course content analysis.

Students' ratings of online courses are direct feedback from their experiences. 
Understanding what elements of online courses may affect students' evaluations and developing an automatic rating prediction system for teachers are of great significance, considering that it can provide practical guidance for teachers to enhance their instructional material design and teaching styles for more effective online teaching in the future.

In our study, we aim to investigate this novel problem of predicting teachers' ratings in online education platforms based on online educational materials. The conceptual illustration of our proposed model is shown in Figure~\ref{fig:concept}. 
In particular, the input of the system is the raw content of the course materials, consisting of the textual subtitles of the lecture videos organized in multiple sections. The output of the system is the predicted student rating of the teacher or the course. We note that there are several non-trivial research challenges in developing such a system due to the special characteristics of course data:

\begin{itemize}
    \item First, teachers' verbal cues may have an important impact on students' evaluation according to traditional education theories. For example, some studies suggest that a teacher who is proficient at asking the question can help students' thinking and interaction \cite{olsher2012asking, clough2007so}. 
    Another research indicates that a speaker who tends to use more hedging words (e.g., a little, kind of, more or less) implies a lack of commitment to the speech content ~\cite{prince1982hedging}, which could be a negative signal of teaching quality. 
    Thus, how to better extract teachers' verbal cues from courses' subtitles using state-of-the-art NLP models entails significant challenges.
    \item Second, every course has a natural hierarchical structure. A course usually consists of multiple sections, and each section has multiple lectures. For example, a course named ``Machine Learning With Python'' has three sections:  Regression, Classification, and Clustering. The Classification section has several different lectures such as K-Nearest Neighbours, Decision Trees, etc. Moreover, the course structure is also an important determinant of the course’s success according to traditional education theory~\cite{bohlin1995course}. Thus, how to leverage course structure information to assist the design of the system for predicting teachers' performance is also challenging.
\end{itemize}

\vpara{Our Contribution.} In our paper, we first conduct a systematic analysis of what elements of online courses may affect students' evaluations. Moreover, we propose a hierarchical course BERT model to accurately predict teachers’ performance in online education using both linguistic features and course structure information. Specifically, we make the following three contributions to this study.

\begin{itemize}
\item First, to the best of our knowledge, we are the first to study the teachers' performance evaluation in online education. Moreover, traditional education studies mainly rely on the self-reported survey and qualitative analysis limited to small-scale analysis. Our study thus contributes to the education literature by conducting a quantitative analysis leveraging machine learning tools using a large-scale dataset.
    
    \item Second, based on traditional education theories, we extract course features including linguistic and course structure characteristics. A model-free analysis shows that question strategy, emotion appealing, hedging, and course structure are significantly correlated with teachers' ratings. These insights can potentially assist teachers to better prepare for online teaching to enhance students' evaluations in the future.

    \item Third, from the methodological perspective, we propose a novel hierarchical course BERT model that can capture courses' natural hierarchical structure information and the deep semantic features from course subtitles.  We demonstrate the superior prediction performance of our approach over several state-of-the-art methods. The proposed approach could be a powerful tool for evaluating and improving teachers' performance in the future, which can benefit millions of students' online learning in our society.
\end{itemize}

\section{Dataset}

We collect a large-scale dataset consisting of 1,085 free online courses on Coursera. Coursera is the largest global platform of MOOCs (Massive Open Online Courses) with more than 40 million learners worldwide and more than 150 partner universities~\cite{coursera}. The free courses on Coursera are available for learners to download. Each course has multiple sections, and each section has multiple lectures. Besides, each course has two ratings displayed on the platform, the course rating and the instructor rating, given by students who have completed the course. The descriptive statistics for these courses are shown in Table~\ref{table:data stats}. On average, each course has 4.95 sections and 40.09 lectures in our data; each section has 8.13 lectures; each lecture has 1158.87 tokens, and most of the lectures are less than 10 minutes. 

\begin{figure}[h]
     \centering

     \begin{subfigure}[b]{0.45\linewidth}
         \centering
         \includegraphics[width=\linewidth]{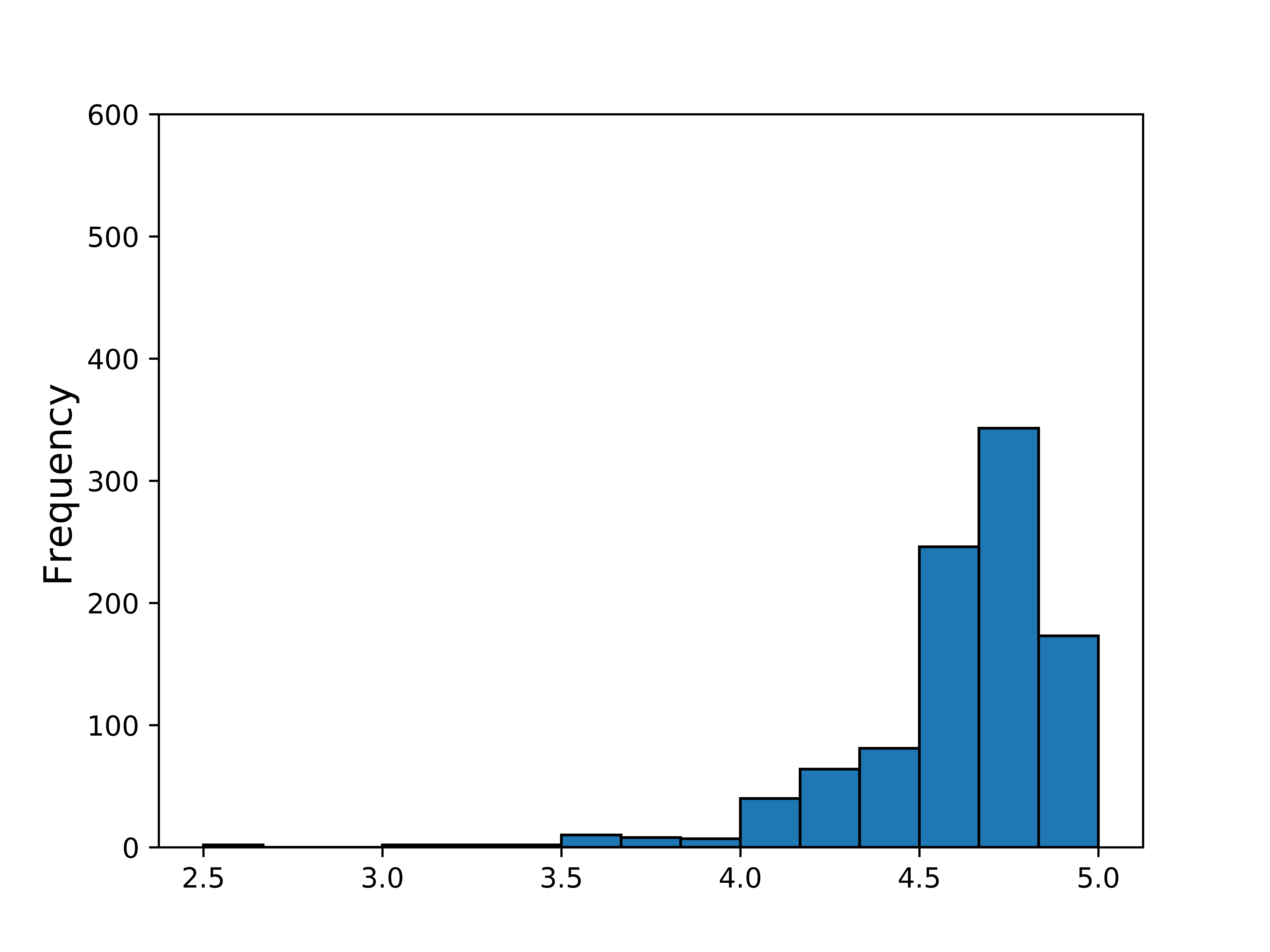}
         \caption{Instructor rating}
     \label{fig:instuctor rating}
     \end{subfigure}
     \hfill
       \begin{subfigure}[b]{0.45\linewidth}
         \centering
         \includegraphics[width=\linewidth]{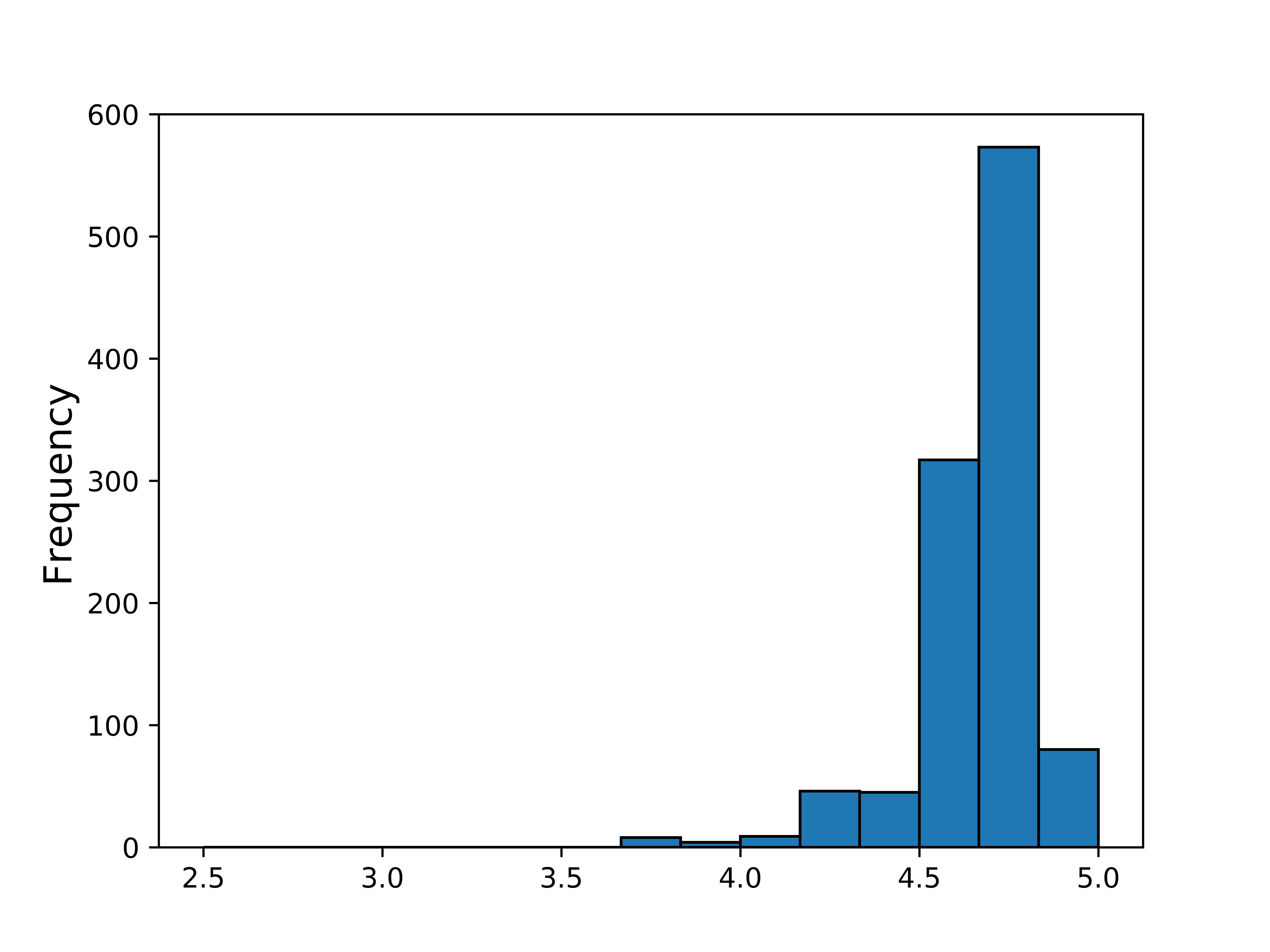}
         \caption{Course rating}
         \label{fig:course rating}
     \end{subfigure} \caption{Distribution of instructor and course ratings}
        \label{fig:rating distribution}
\end{figure}

The course rating is aggregated from students’ feedback collected when they complete the course. For the instructor rating, the platform asks all learners to provide feedback for instructors based on the quality of their teaching style. It should be noted that the instructor with multiple courses has multiple ratings rated by the students who have taken that specific course. The distribution of the two ratings is provided in Figure~\ref{fig:rating distribution}. Both ratings range from 0.0 to 5.0. 

\begin{table}[h]
	\centering
	 \caption{Descriptive statistics for online courses}
	\begin{tabular}{lrrrr}
	\toprule
	  & Mean & SD & Median\\
\midrule
\# Sections & 4.95&1.81&4.00 \\
\# Lectures/course&40.09&25.01&35.00 \\
\# Lectures/section&8.13&5.24&7.00 \\
\# Tokens/lecture&1158.87&855.50&969.00\\
\bottomrule
	\end{tabular}
	\label{table:data stats}
\end{table}

\section{Exploration of Rating-Correlated Signals}
\label{sec:features}

We have conducted a series of exploratory studies to find out potential strong indicators for higher or lower ratings. 


\subsection{Extracted Course Features}

Based on traditional education literature, we extract two sets of course features in our data -- teachers' verbal cues and meta-information features.
\subsubsection{Part 1. Teachers' Verbal Cues.}

Six linguistic features are extracted by gathering existing hand-crafted, linguistic lexicons for relevant concepts.
\begin{itemize}
    \item \textbf{Concreteness Ratio.}   Concreteness of the course content may have an important impact on students' comprehension, interest, and learning outcomes~\cite{sadoski2000engaging}. A larger concreteness ratio represents very concentrated teaching content and less unrelated materials. To quantify the concreteness ratio, we first count the total number of named entities for each course in our data.
It has five coarse-grained groups: (1) events, (2) numbers, (3)
organizations/locations, (4) persons, and (5) products. Then, a concreteness ratio, namely, the number of named entities in the course divided by the total number of tokens in that course, is calculated. 

\item \textbf{Questions Ratio.} Asking  questions in class can help students’ thinking and interaction~\cite{olsher2012asking, clough2007so}. To quantify teachers' question-asking behavior in our data, we use a question detection method -- Basic Parse Tree created by Stanford's CORE NLP\footnote{https://stanfordnlp.github.io/CoreNLP/} for each sentence to detect if the sentence is a question. We then calculate the question ratio,  defined as the number of question divided by the  number of sentences in each course. We use the Penn Treebank’s Clause level tags for question detection. We specifically check for the occurrence of two tags: (1) \textit{SBARQ} -- direct question introduced by a wh-word or a wh-phrase and (2) \textit{SQ} -- inverted yes/no questions.



\item \textbf{Emotion Appealing.} This refers to teachers' elicitation of emotions which may attract students' attention while learning~\cite{wang2019persuasion}. NRC Sentiment Lexicons~\cite{mohammad2013nrc} is used to extract each teacher’s emotions along eight dimensions: anticipation, joy, surprise, trust, anger, disgust, fear, and sadness. Then, the emotion entropy is calculated to measure the emotion diversity of a teacher, which is defined as $-\sum_{i=1}^{8} r_i \times log(r_i)$ where $r_i$ denotes the ratio of emotion tokens to total tokens for each specific emotion dimension  $i$.

\item \textbf{Hedging.}
Hedges are lexical choices made by a speaker, which may indicate a lack of commitment to the content of their speech ~\cite{prince1982hedging}. We use the single- and multi-word hedge lexicons
from~\cite{prokofieva2014hedging} to calculate the ratio of hedges to tokens in each course. For example, teachers who tend to use ``basically, generally, sometimes (unigrams)'', ``a little, kind of, more or less (multi-word)'' may imply that they are not quite confident about their teaching materials, which might be a signal of poor teaching quality.

\item \textbf{Other Lexicon-Based Features.}
We also compute several other lexicon-based features, including ratios of (1) strong modal such as "always", "clearly", "undoubtedly" and (2) weak modal such as  "appears", "could", "possibly" using the respective lexicons from~\cite{loughran2011liability}. In each case, the ratio of terms in the respective category to the number of tokens in each course is computed.
\end{itemize}

\subsubsection{Part 2. Meta-Information Features.}
In the second set of our features, we compute several measurements on course-level meta-information.
\begin{itemize}
    \item \textbf{Course Length.}
The total number of tokens in each course is calculated to represent the course length. A longer course could be an indicator of more elaborate teaching which may have an impact on students~\cite{ladson1992culturally}.
\item \textbf{Course Structure Quality}. The structure of the course is also a  crucial factor for online education~\cite{bohlin1995course}. Each course has a default structure, namely, from course to sections, and from sections to lectures. For a well-structured course, every lecture within the same sections should be highly connected but less similar to the lecture from other sections. We use modularity -- the measure of the structure of networks or graphs -- to measure the quality of the course structure in our data. 
In our study, a lecture network, where each course has a network and nodes are individual lectures, is established. The edge is calculated by the pair-wise cosine similarity of Word2Vec embedding of each lecture. Given the lecture network $G$ and the current default structure $S$, the modularity is calculated to measure the quality of the current default structure as follows:

\begin{equation}
        Q = \frac{1}{2m} \sum_{ij} \left( A_{ij} - \frac{k_ik_j}{2m}\right)
        \delta(c_i,c_j)
\end{equation}
where $m$ is the number of edges, $A$ is the adjacency matrix of
$G$, $k_i$ is the degree of $i$ and $\delta(c_i, c_j)$
is 1 if $i$ and $j$ are in the same community and 0 otherwise. The graph building and modularity calculation are implemented using NetworkX\footnote{https://networkx.github.io/documentation/stable/tutorial.html}.
Particularly, the higher the modularity, the better the course structure quality.
\end{itemize}


\subsection{Interpretation of Correlation Results}

We present the results for the correlation analysis in Table~\ref{table:corr}. In particular, we find some interesting patterns that may guide teachers in an online education setting to enhance students’ evaluations in the future. We expand the statistically significant results for negative ($-$) and positive ($+$) correlations with the instructor rating.

\begin{itemize}
    \item ($+$) Question ratio. This suggests that
the question-asking strategy may enhance students' learning experience, regardless of the lack of direct interaction with students in the online education setting. Because it can potentially arouse students’ thinking and attract students’ attention.

     \item ($+$) Emotion appealing.  This indicates that the entertainment value of a course is also crucial, and it is consistent with online consumer experience theory in the marketing literature~\cite{woltman2003consumers}. In a traditional course in the physical classroom, the teacher and students can have direct interactions and eye-contact, and students are required to sit in the classroom until the class ends. However, in the online education setting, students can leave the course whenever they want and it becomes even more challenging for the teacher to engage students in a class. Teachers who tend to use an emotion appealing strategy could potentially enhance students' experience and better engage students in the online class.   
     
       \item ($-$) Hedging. This refers to those teachers who tend to use more hedging words such as ``a little'', ``kind of'', and ``more or less'' might be less committed to the content of their speech. Therefore, a higher percentage of hedging words might be a negative signal of teaching quality.
    
      \item ($+$) Course structure quality. This indicates that the structure quality of the online course might be an essential factor in students’ online learning. A well-structured course is an indicator of a teacher’s effort in preparing the course and its potential high quality.
      
            \item ($+$) Course length.  
            The length of a course is an indicator of the amount of content contained in the course. A longer course may enable the students to acquire more knowledge from the course and thus result in more positive evaluations.
\end{itemize}

It is worth noting that our study only reveals the correlations between these signals and the ratings, and this study aims to build a better rating prediction system. However, we caution that strong correlations do not necessarily imply causality.

\begin{table}[h!]
	\centering
	 \caption{Results from Pearson correlations between extracted course features  and  ratings. Statistical
significance is marked by
(*) for p $\textless$ 0.05. }
	\begin{tabular}{lrrrr}
	\toprule
	\multirow{2}{*}{Features} & \multicolumn{2}{c}{Instructor rating} & \multicolumn{2}{c}{Course rating}\\

	  & $Corr$  & $p$-value & $Corr$  & $p$-value  \\
\midrule
Concreteness &0.02&0.48 &0.01&0.70 \\
Questions & 0.48* &\textless $10^{-2}$&0.04*&0.03 \\ 
Emotion appealing &0.16* & \textless $10^{-4}$&0.17*&\textless $10^{-4}$\\
Hedging  &-0.06*& 0.03&-0.06*&0.03\\ 
Strong modal &-0.12* &\textless $10^{-4}$&-0.11*
&\textless $10^{-2}$\\

Weak modal &0.08* &\textless $10^{-2}$ &0.04&0.14\\

Course length &0.18* &\textless $10^{-4}$&0.26*&\textless $10^{-4}$\\
Course structure &0.06*&0.04 &0.10*&\textless $10^{-2}$\\
\bottomrule
	\end{tabular}
	\label{table:corr}
\end{table}

\section{Hierarchical Course BERT Model for Teachers' Performance Prediction}

As demonstrated in the previous section, teachers’ verbal cues and course structure quality are significantly correlated with the ratings. This motivates us to build a more advanced model to better capture teachers’ verbal cues and course structure quality in each course. Pretrained language models (e.g., BERT~\cite{devlin2018bert,beltagy2020longformer}) have achieved state-of-the-art performance on a wide range of downstream natural language understanding (NLU) tasks, such as the GLUE benchmark~\cite{wang2018glue} and SQuAD~\cite{rajpurkar2016squad}.

\begin{figure}[tb]
    \centering
    \includegraphics[width=0.8\linewidth]{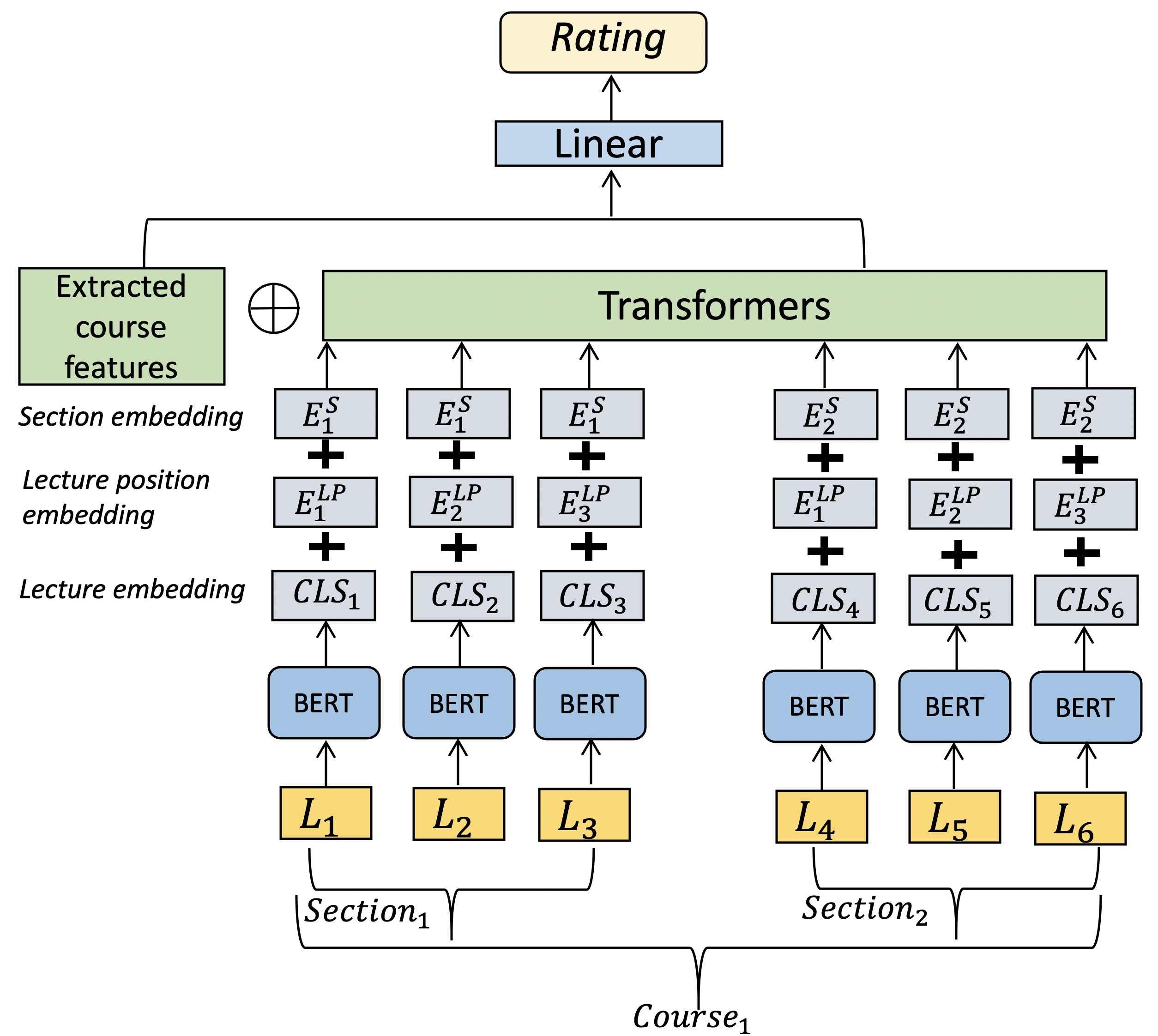}
    \caption{The structure of our proposed hierarchical course BERT model for teachers' performance prediction. $L$ refers to the lecture. }
    \label{fig:bert}
\end{figure}

In our study, we propose a hierarchical course BERT model to better capture the course structure quality and linguistic features in each course. The overall structure is shown in Figure~\ref{fig:bert}. Assuming that $\{\mathbf x_i, \mathbf s_i, \bpi_i, \mathbf p_i, y_i\}_{i=1}^N$ represent $N$ courses, where $\mathbf x_i = \{\mathbf l_{i1},\mathbf l_{i2},...,\mathbf l_{iJ_i}\}$ is $i_{th}$ course subtitles with $J_i$ lectures $\mathbf l$. Each lecture $\mathbf l_{ij}$ is a subtitle, represented by a sequence of tokens. $\mathbf s_i = \{s_{i1}, s_{i1}, ...,s_{iJ_i}\}$ contains $J_i$ integers $1 \leq s_{ij} \leq N$ denoting the section index of each lecture to indicate which section the lecture belongs to; $\bpi_i = \{\pi_{i1}, \pi_{i2}, ..., \pi_{iJ_i}\}$ contains integers denoting the relative lecture position within each section; $\mathbf p_i$ is an additional feature vector for course $i$ containing numerical features extracted in the section above.
$y_i$ is the corresponding ground-truth label (i.e., ratings).

\vpara{Local lecture semantic features.} 
For course $i$ with $J_i$ lectures, we cannot combine all the lectures' tokens and feed them through a BERT model due to its 512-token length limit. Since each lecture has 1158.87 tokens on average, we take each individual lecture subtitle as a unit. For  lecture $j$ where the subtitle sequence is denoted as $\mathbf l_{ij}$, we add a special token $[CLS]$ at the beginning of each lecture, and do a forward pass in a pretrained BERT model. Then we take the $[CLS]$ output embedding  $\mathbf h_{ij} \in \mathbb{R} ^{H}$ for lecture $j$ to represent the lecture, where $H$ is hidden size (e.g., 768 hidden units).

\vpara{Incorporating course structure.}  \cite{bai2020segabert} proposed a segment-aware BERT  by adding additional paragraph index,
sentence index, and token index embeddings to encode the natural document's hierarchical structure. Compared to the original BERT model on various NLP tasks, it exhibits a significant gain on predictive performance. Motivated by their findings, in our model additional section embedding is added to encode the lecture’s section position information (i.e., which section does each lecture belong to), and lecture position embedding to encode the order of lectures within each section on top of the lectures $[CLS]$ token sequence extracted from lecture subtitles.  More specifically, for each course, all the lecture $[CLS]$ tokens in the same section (e.g., section $1$) are assigned the same section embedding $E^S_{1}$.  Within each section, we use the lecture position to encode the order of lectures in each section. For example the first lecture in every section would be assigned the same lecture position embedding $E^{LP}_{1}$. 
Besides, the maximum section number is set to 8 since 95\% of the courses in our data have less than 8 sections.  Similarly, the maximum lecture positions within each section are set to 10, since 90\% of the courses in our data have less than 10 sections. 

We add three embedding together: (1) the lecture embedding which captures the semantic features of each lecture, (2) section embedding, and (3) lecture position embedding to obtain  the lecture representation $\mathbf z_{ij}$:
\begin{align}
    \mathbf z_{ij} = \mathbf h_{ij} + E^{S}_{s_{ij}} + E^{LP}_{\pi_{ij}}
\end{align}

\vpara{Global transformers to exchange the lectures information.} 
We take the sequence of all lectures' representation $(\mathbf z_{i1}, \mathbf z_{i2}, ..., \mathbf z_{iJ_i})$ and 
add a new special token -- course $[CLS]$ at the beginning to represent the whole course, denoted as $\mathbf z_i^0$.

We use the global transformers to exchange the lectures tokens information within same course.  Here we use vanilla transformer layer~\cite{vaswani2017attention} and composed of two sub-layers.
\begin{align}
    \mathbf h^l_i &= \text{LayerNorm}(\mathbf z^{l-1}_i+\text{MHAtt}(\mathbf z^{l-1}_i)) \\
    \mathbf z^l_i &= \text{LayerNorm}(\mathbf h^l_i +\text{FFN}(\mathbf h^l_i))
\end{align}
where LayerNorm is a layer normalization proposed in \cite{ba2016layer}; MHAtt is the multihead attention mechanism introduced in ~\cite{vaswani2017attention} which allows each token to attend
to other tokens with different attention distributions; and FFN is a two-layer feed-forward network with ReLU as the activation function.

We take the course $[CLS]$ token representation
output by the last layer of the global transformers to represent the course semantic  and structural features, denoted as $\mathbf z_i$.

\vpara{Extracted course features.} In addition to the course $[CLS]$ tokens,  extracted course features $\mathbf p_i$ shown in the previous correlation sections are also concatenated. These features are different from the semantic features extracted from BERT, since they are extracted using human knowledge, such as the emotion appealing and questions detection.

\vpara{Linear layer.} We  concatenate course $[CLS]$ token $\mathbf z_i$  and extracted  course features $\mathbf p_i$ together as $\mathbf z_i \oplus \mathbf p_i$. Then, the concatenated features are put on a linear layer to conduct the final rating prediction $\hat{y}_i$.

\begin{equation}
  \hat{y}_i = \mathbf W[\mathbf z_i \oplus \mathbf p_i] + b
\end{equation}
where $W$ denotes weight matrix and $b$ denotes bias.

\section{Experiments}

\subsection{Experiment Setup}
We experiment with two prediction tasks -- instructor rating and course rating prediction.
We split 70\%, 10\%, and 20\% of courses as training, validation, and test set. 
We minimize the MSE loss for training. We take pretrained BERT$_{base}$ with 12 layers to encode local semantic features from each lecture. Pretrained
model weights are obtained from Pytorch transformer repository\footnote{https://github.com/huggingface/PyTorch-transformers}. We also update the BERT$_{base}$ parameters during training. The global transformer layers are trained from scratch. Besides, we set the number of global transformer layers as 2 based on the preliminary experiments. We find the 2 layer global transformer layers work much better than 1 layer. Due to the resource limitation, we did not try three layers, and we leave this to future work. We set the number of the final linear layer as 1.  All transformer-based models/layers have 768 hidden units.  The model is trained on Nvidia V100 GPU. The
optimizer is Adam~\cite{kingma2014adam} with
learning rate of 3e-5, $\beta_1$ = 0.9, and $\beta_2$ = 0.998;
we also applied learning rate warmup over the
first 500 steps, and decay as in~\cite{vaswani2017attention}.  We set the epochs as 50,  batch size as 16. We apply dropout (with a probability of 0.1) before all linear layers. 


\subsection{Baselines}
Our proposed hierarchical course BERT model is compared with various baselines.

\textit{(1) Extracted course features only.} Since extracted course features are part of our proposed hierarchical course BERT model. If the extracted course features have excellent prediction performance, there is no need to build an advanced model. The regression with extracted course features is conducted using seven machine learning algorithms to control for the variability, including linear regression, support vector machines with the radial based kernel (SVM), Multi-layer Perceptron (MLP), Random Forest, AdaBoost, GradientBoosting, and Bagging.  For the ensemble method like AdaBoost and Bagging, we use a decision tree as a base component in the ensemble. We perform a grid search in $\{10,50,100\}$ to find the optimal number of components for ensemble methods on the validation set. For MLP, we perform a grid search to find the optimal number of hidden layers ($H_l$) and hidden size ($H_s$) ($H_l \in \{1,2,3\}$, $H_s \in \{20, 50, 80, 100\}$). Adam optimizer is used to optimize the parameters.

\textit{(2) Word2Vec + extracted course features.}
The pre-trained word2vec model on the entire Google News dataset is adopted to obtain 300-dimensional word embedding.  Besides, all words in the same course are added to represent the course. These semantic features are similar to BERT features without considering structure information. Moreover, the extracted course features are concatenated to the course word2vec embedding. We repeat the same machine learning algorithms and the hyperparameter tuning process as (1).

\textit{(3) Doc2Vec + extracted course features.}
The paragraph vector algorithm
proposed by~\cite{le2014distributed} is used to obtain
300-dimensional lecture embeddings. Using the Gensim implementation ~\cite{rehurek2010software}, doc2vec models for 50 epochs are trained while words occurring less than 10 times in the respective training corpus are ignored. Doc2vec embeddings are trained over individuals lectures. Besides, we add the doc2vec embedding for lectures in the same course together as the course embedding features. Then, the extracted course features are concatenated to the course doc2vec embedding. We repeat the same machine learning algorithms and the hyperparameter tuning process as (1).

\textit{(4) Lecture BERT (LecBERT).} In this baseline model, each lecture is regarded as a unit.  This baseline does not have access to the course structure information. We assume lectures in the same course share the same rating. We assign the same rating score to the lectures in the same course for training. For each lecture, the lecture $[CLS]$ is obtained after the BERT model, and the linear layer directly is added after the BERT model to make lecture rating prediction. In the test stage, prediction for each lecture is first conducted; then, we take the average over all lectures in the same course to obtain the course rating. The pretrained BERT$_{base}$ with 12 layers to encode semantic features from each lecture. We set the epochs as 5,  batch size as 64. Other hyperparameters are same as proposed model.



\subsection{Performance Metrics}
We use the following two metrics to measure the prediction performance:

\begin{itemize}
    \item \textbf{Root mean squared error (RMSE)}:
    \begin{equation}
        \text{RMSE} = \sqrt{\frac{1}{N}\sum_{j=1}^N (y_j - \hat{y}_j)^2}
    \end{equation}
    \item \textbf{Mean Absolute Error (MAE)}:
        \begin{equation}
        \text{MAE} = \frac{1}{N}\sum_{j=1}^N |y_j - \hat{y}_j|
    \end{equation}
\end{itemize}

\subsection{Results}
\vpara{Overall comparison.} The results are presented in Table~\ref{table:experiment results}. Our proposed model performs the best among all the baselines for both the instructor rating prediction task and the course rating prediction task. GradientBoosting in the group of Dov2vec + course is the best baseline model among the course features alone group, word2vec+course group, and doc2vec+course group. Compared to the best baseline model, our proposed model reduces  RMSE by 49.6\% ($p$ \textless 0.01) and MAE by 63.8\% ($p$\textless0.01) for the instructor rating task. Regarding the course rating task, our proposed model reduces the RMSE by 45.7\% ($p$\textless0.01) and MAE by 62.0\% ($p$\textless0.01) compared to the best baseline model. 

 Besides, compared to course features only, adding additional semantic features (e.g., Word2vec or Doc2vec) can largely increase the performance for both instructor rating prediction task and course rating prediction task. For example, with Bagging, the group of Doc2vec + course reduces RMSE by 12.4\% and MAE by 16.2\% compared to the course features alone for instructor rating tasks. Also, adding Doc2vec embedding performs better than Word2Vec. This indicates (1) better semantic features and (2) combination of deep semantic features and extracted course features are two ways to get better performance. Our proposed model builds on these findings.

\begin{table}[t!]
\small
	\centering
	 \caption{  Experiment results. LR refers to Linear Regression. RF refers to Random Forest. GB refers to GradientBoosting.
	 Our proposed model works the best on both instructor and course prediction task. }
	\begin{tabular}{p{1.1cm}rrrrrr}
	\toprule
		\multirow{2}{*}{Class} &\multirow{2}{*}{Model}& \multicolumn{2}{c}{Instructor rating} & \multicolumn{2}{c}{Course rating}\\
 &  & RMSE  & MAE & RMSE  & MAE \\
\midrule
\multirow{7}{*}{\parbox{1.1cm}{Extracted course features only}} & LR &0.8725 &0.6901&0.4075&0.3571 \\
&SVM&0.3001&0.2149&0.1809	&0.1324\\
&MLP& 0.3079& 0.2202&0.1834&	0.1427\\
 &RF&0.3206 & 0.2351&0.1886&	0.1402\\
&AdaBoost&0.3812 & 0.3302&0.2093&	0.1750\\
&GB& 0.3017& 0.2161&0.1823	&0.1325\\
&Bagging& 0.3159&0.2348 &0.1847&	0.1392 \\
\hline %
\multirow{7}{*}{\parbox{1.1cm}{Word2vec + \\ course}} & LR & 0.3859&0.2790 &0.2475&	0.1994 \\
&SVM&0.2840& 0.2063&0.1803&	0.1323\\
&MLP&0.2966 &0.2189&0.1786	&0.1368\\
&RF&0.2881 & 0.2095&0.1773	&0.1351\\
&AdaBoost&0.2774 & 0.2130&0.1766&0.1393\\
&GB&0.2873 &0.2077&0.1773&	0.1313 \\
&Bagging& 0.2963&0.2157&0.1788&	0.1349 \\
\hline %
\multirow{7}{*}{\parbox{1.1cm}{Doc2vec + \\ course}} & LR &0.3149 & 0.2223&0.2082	&0.1604\\
&SVM&0.2809&0.2018&0.1727&	0.1319 \\
&MLP& 0.2920& 0.2170 &0.1776&	0.1380\\
&RF&0.2756 &0.1959&0.1744&	0.1265\\
&AdaBoost& 0.2739& 0.2107&0.1710&0.1319\\
&GB& 0.2739& 0.1939&0.1680	&0.1238\\
&Bagging& 0.2767&0.1966&0.1709	&0.1244 \\
\hline %

\multirow{2}{*}{\parbox{1.1cm}{BERT Variant}} & LecBERT & 0.2432& 0.1615&0.1453&	0.1153\\
&\textbf{Our model}& \textbf{0.1378}& \textbf{0.0701}&\textbf{0.0913}&	\textbf{0.0471}\\
\bottomrule
	\end{tabular}
	\label{table:experiment results}
\end{table}

\vpara{Ablation study.} The importance of each element in our proposed model is investigated. Then, the following three models are further conducted by removing items step by step.

\textit{(1) StrucCourse BERT.} The concatenated course features to the model are removed, and other specification from our proposed model is maintained, to study the importance of the sparse feature in addition to the dense features.

\textit{(2) Course BERT.} For this model, the structure information is further removed from StrucCourse BERT (i.e., section embedding and lecture position embedding). However, we keep the global transformers to exchange the lecture information, suggesting that we drop the structure information while keeping the lectures information exchange. For this model, each lecture is first fed into the BERT model; then, the lecture $[CLS]$ tokens are directly put into transformers without adding the section embedding and lecture position embedding to explore the importance of course structure information.

\textit{(3) Lecture BERT.} The lecture exchange information is further removed from Course BERT. Specifically, we remove the global transformer layers and treat each lecture as an independent.

\begin{figure}[tb]
    \centering
    \includegraphics[width=0.9\linewidth]{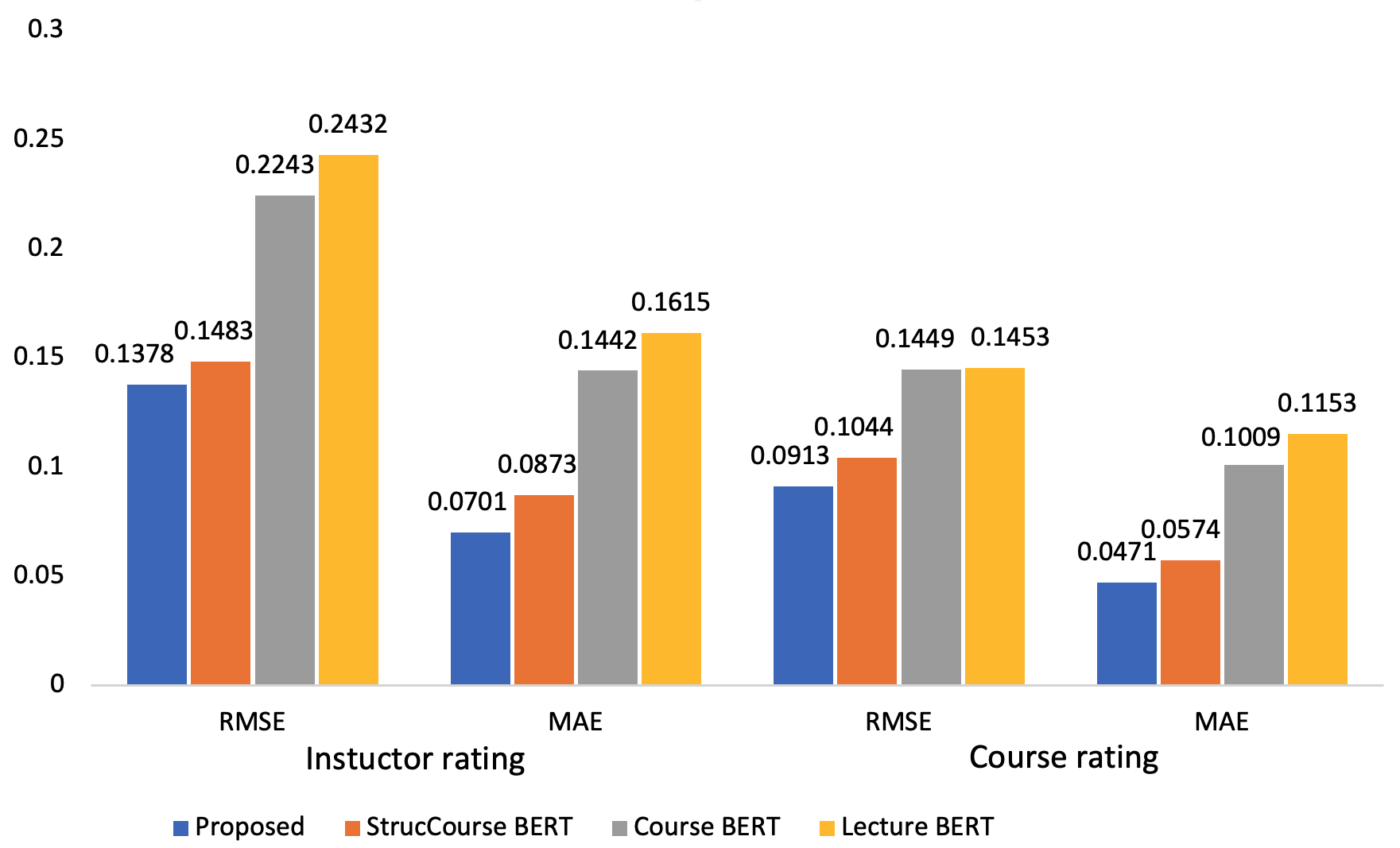}
    \caption{Ablation study on the importance of each element in the proposed model. The course structure information provides significant gain (- 33.9\% RMSE and -39.5\% MAE on instructor rating task) and extracted course features can provide some additional gain (- 7.1\% RMSE and -19.7\% MAE on instructor rating task).}
    \label{fig:ablation}
\end{figure}

The results are summarized in Figure~\ref{fig:ablation}. We find the course structure information is the most important element in our model. The course sparse features and global transformers provide additional gains.
Some interesting findings are highlighted as follows.
\begin{itemize}
    \item \textbf{Extracted course features can further improve the performance.} Adding the extracted course features can further improve the performance. Compared to StrucCourse BERT, the model with additional extracted course features can provide additional gains. Regarding the instructor rating prediction task, our proposed model reduces RMSE by 7.1\% and MAE by 19.7\% compared to StrucCourse BERT. Regarding the course rating prediction task, our proposed model reduces RMSE by 12.5\% and MAE by 17.9\% compared to StrucCourse BERT.
    
    \item \textbf{Course structure information provides a significant gain.} Adding section embedding and lecture position embedding to represent the natural course structure can significantly improve the performance. Compared to the Course BERT, StrucCourse BERT with additional section embedding and lecture position embedding reduces RMSE by 33.9\% and MAE by 39.5\% for the instructor rating prediction task. For the course rating prediction task, StrucCourse BERT reduces RMSE by 28.0\% and MAE by 43.1\% compared to the Course BERT.
    
    \item \textbf{Global transformers to exchanging the lecture information are important.}  Compared to Lecture BERT treating lectures as independent, the Course BERT using global transformers to exchange the lectures information within the same course exhibits better prediction performance. For example, regarding the instructor rating prediction task, the Course BERT reduces RMSE by 7.7\% and MAE by 10.7\% compared to Lecture BERT; regarding the course rating prediction task, the Course BERT reduces MAE by 12.5\% compared to Lecture BERT. 

\end{itemize}



\section{Related Work}

\vpara{Online education mining.} Our work focuses on education mining broadly. There are some MOOC-related studies on student behavior analysis~\cite{feng2019understanding,trakunphutthirak2019study}, course concept mining~\cite{pan2017course,yu2019course}, and course recommendation~\cite{zhang2019hierarchical}. In our paper, we have addressed an important research gap in the existing literature on what makes a star teacher in online education by analyzing teachers’ verbal cues and course structure. Based on our exploratory insights, we further propose a novel hierarchical BERT model to evaluate teachers’ performance in online education. Our study contributes to the literature by providing a practical tool that can be used by teachers to enhance their instructional materials for more effective online teaching in the future.



\vpara{Traditional education theory.}  Our work is also closely related to traditional education studies (e.g., teacher’s performance evaluation ~\cite{gordon2006identifying,deci1982effects,rowan1997using} and teaching skills~\cite{shavelson1973basic, van2014teaching}). However, these studies mainly depend on the survey or qualitative methods limited to small-scale data. We contribute to the literature by quantifying teachers’ verbal cues and teaching strategy from detailed teaching subtitles and course structure using scalable and automatic methods. Moreover, our proposed model can be used to accurately predict teachers’ performance to help teachers better prepare for online teaching in the future.

\vpara{BERT-based real-world applications.} Prior research applying BERT to various real-world problems have achieved huge success in many tasks such as humour detection, hate speech detection, fake news detection, and adverse drug reaction detection from tweets~\cite{mao2019bert,liu2019two,mozafari2019bert, roitero2020twitter,breden2020detecting}. To the best of our knowledge, our paper is the first study to apply the BERT model to course corpus and achieve a significant performance result.

\section{Conclusion and Future Work}

In this paper, we conduct a systematic study to evaluate teachers’ performance in online education using courses' subtitles. Our results indicate that teachers’ verbal cues and course structure quality have significant correlations with teachers' performance in online education. Based on the gained insights, we propose a novel hierarchical course BERT model to accurately predict teachers’ performance. Our proposed model can capture the course’s natural hierarchical structure and the deep semantic features from course subtitles. Our experiments demonstrate that our proposed method achieves a significant gain over several state-of-the-art methods. Our study provides immediate and actionable implications for teachers by developing a practical tool for teachers to predict and enhance their teaching performance for more effective online education in the future. For future work, 
multi-modal features (e.g., audio and visual features) can be extracted to have  a more robust model.
\bibliography{ref}

\end{document}